\documentclass{article}
\usepackage{PRIMEarxiv}

\usepackage[english]{babel}

\usepackage{amsmath}
\usepackage{graphicx}
\usepackage[colorlinks=true, allcolors=blue]{hyperref}
\usepackage{url} 
\title{Bias in Text Embedding Models}
\author{Vasyl Rakivnenko, Nestor Maslej, Jessica Cervi and Volodymyr Zhukov}

\begin{document}
\maketitle

\begin{abstract}
Text embedding is becoming an increasingly popular AI methodology, especially among businesses, yet the potential of text embedding models to be biased is not well understood. This paper examines the degree to which a selection of popular text embedding models are biased, particularly along gendered dimensions. More specifically, this paper studies the degree to which these models associate a list of given professions with gendered terms. The analysis reveals that text embedding models are prone to gendered biases but in varying ways. Although there are certain inter-model commonalities, for instance, greater association of professions like nurse, homemaker, and socialite with female identifiers, and greater association of professions like CEO, manager, and boss with male identifiers, not all models make the same gendered associations for each occupation. Furthermore, the magnitude and directionality of bias can also vary on a model-by-model basis and depend on the particular words models are prompted with. This paper demonstrates that gender bias afflicts text embedding models and suggests that businesses using this technology need to be mindful of the specific dimensions of this problem.
\end{abstract}

\section{Introduction} 
Text embedding is an efficient AI approach that works to transform text into numerical representations ~\cite{cer2018universal, devlin2018bert, mikolov2013efficient}. Text embedding has a plethora of potential business applications ~\cite{su2023learning}. Despite the growing popularity of AI and AI text-embedding systems, the degree to which these systems are biased is not as well-understood. This lack of understanding persists despite the fact that it is now fairly well-documented that AI systems more broadly, are quite prone to bias ~\cite{buolamwini2018gender, larrazabal2020gender, levin2016beauty, maslej2023artificial, mehrabi2021survey, zhang2022ai}. 

\noindent This paper studies bias, in particular gender bias, in popular text embedding models. The analysis proceeds by selecting a list of popular occupations (for example nurse, CEO, writer), and then examining the degree to which a selection of popular text embedding models (selected from an assortment of leading AI developers) associates each of those occupations with gendered terms like boy, girl, woman and man. The text embedding models studied include: (1) \texttt{AI21-v1-embed}, (2) \texttt{amazon-titan-embed-text-v1}, (3) \texttt{baai-bge-large-zh-v1.5}, (4) \texttt{cohere-embed-english-v3.0}, (5) \texttt{bert-large}, (6) \texttt{llama-2-70b}, (7) \texttt{msmarco-distilbert-cos-v5}, (8) \texttt{openai-text-embedding-ada002} and (9) \texttt{voyageai-voyage-01}.

\noindent Each of the text embedding models featured in the analysis routinely associates certain professions with gendered terms, thereby demonstrating some degree of gendered bias. While there are certain cross-model commonalities, for instance, homemaking and caring professions like nurse and homemaker are more strongly associated with female identifiers, and leadership positions are more strongly associated with male identifiers, not all models make the same types of gendered associations. The magnitude of gendered associations also varies strongly across models: some models are shown to be substantially more biased than others. Finally, the gendered associations models make can be quite sensitive to the particular words models are prompted with. 

\noindent All in all, this analysis suggests that gender bias is a somewhat idiosyncratic problem in text embedding models. Businesses that build with these models need to be broadly mindful of the problem of bias and specifically consider the interactions between the text-embedding models and the word combinations they use.

\section{Motivation}

\subsection{Problem Definition}

\noindent Consider the following riddle: A man and his son get into a terrible car crash. The father dies, and the boy is badly injured. In the hospital, the surgeon looks at the patient and exclaims, ``I cannot operate on this boy, he is my son. Who is the surgeon?" A majority of people are reportedly unable to solve this riddle, and this fact is taken as evidence of the widespread prevalence of implicit gender bias ~\cite{belle2021comelongway}. Many people who hear the riddle for the first time have difficulty assigning both the role of ``mother" and ``surgeon" to the same person ~\cite{belle2021comelongway}.

\subsection{Bias in AI}

There are many ways in which AI systems could be biased. Bias can be defined as a ``systematic error in decision-making processes that results in unfair outcomes," and in the context of AI, bias can result from "data collection, algorithm design, and human interpretation" \cite{ferrara2023fairness}.

\noindent There are many types of AI bias ~\cite{ferrara2023fairness}. For example, there is sampling bias, which arises when an algorithm's training data is not fully representative of the population on which it is ultimately deployed. There is now widespread evidence that many leading facial recognition algorithms, which are trained mostly on datasets of white faces, are less accurate at recognizing individuals with darker skin tones ~\cite{buolamwini2018gender, khalil2020systematic, levin2016beauty}. There have also been documented instances of medical image classification systems, which were predominantly trained on images of white adult males, degrading in performance when asked to classify the images of non-white population groups ~\cite{larrazabal2020gender, mittermaier2023bias, seyyed2021underdiagnosis}. 

\noindent Algorithmic bias is another form of AI bias where algorithms are designed in such a way to prefer certain features, which can lead to unfair selection outcomes. A notable example of algorithmic bias occurred when Amazon created an AI-hiring tool trained predominantly on a mostly male dataset of historically successful Amazon applicants ~\cite{dastin2018bias}. As such, the algorithm associated hiring success with being male and would punish applicants if their applications included references to gender (for example, an applicant participating in an all-women's choir). Overall, it is now fairly well-established that AI systems, if improperly designed, can be biased ~\cite{maslej2023artificial, mehrabi2021survey, zhang2022ai}.

\subsection{Vector Embeddings and Bias}

Embedding is an efficient approach for feature extraction in the fields of data science and natural language processing (NLP) ~\cite{cer2018universal, devlin2018bert, mikolov2013efficient}. Machine learning (ML) algorithms cannot work directly with text or symbols. The text such systems interact with needs to be transformed into numerical representations. Embedding is thereby the process of representing words or phrases in a high-dimensional numerical vector space. Embeddings can be powerful because they are less sensitive to misspellings, synonyms, and phrasing differences ~\cite{su2023learning}. They can even work in cross-lingual scenarios ~\cite{su2023learning}.

\noindent Vector embedding models like Google's \texttt{bert}, Cohere's \texttt{embed}, and OpenAI's \texttt{ada-002} can be deployed in a variety of data-centered systems ~\cite{cohere2024embed, devlin2018bert, openAI2022embed}. For instance, Google's \texttt{bert} is used to parse the search intentions of users, Cohere's \texttt{embed} is an underlying technology for their Classify model, and OpenAI's \texttt{ada-002} is a crucial backbone of Microsoft Azure's cognitive search ~\cite{cohere2024classify, nayak2019bertsearch, microsoft2024azure}. An important feature in the success of embedding models are powerful embedding layers, which learn continuous representations of input information such as words, sentences, and documents from large amounts of data. If irrelevant content is passed to the embedding model, it counters the objective of efficient embedding and requires models to filter extraneous information. These types of inefficiencies can reduce the quality of generated responses, increase latency, and thereby increase operating costs ~\cite{harris2024enhancing, muennighoff2022mteb, wang2019evaluating}.

\noindent As noted earlier, there is a plethora of scholarship that has established that AI systems can be biased. As LLMs have become increasingly popular in recent years, there has also been emerging scholarship highlighting how LLMs can be biased. It is now well-understood that LLMs like GPT-3.5 and GPT-4 can exhibit bias, especially along gender dimensions ~\cite{kotek2023gender}, that LLMs can more broadly absorb biases from cultural associations present in language corpora ~\cite{caliskan2017semantics}, and that the bias of such models can have fairly wide-ranging downstream implications ~\cite{barocas2023fairness, rudinger2018gender}. Scholars have also discussed bias mitigating strategies for LLMs ~\cite{gonen2019lipstick, su2023learning, thakur2023language}.

\noindent However, in the existing literature, there are not many systematic comparisons of the relative bias of various language models, especially text embedding models. In other words, there are few current studies that ascertain which text embedding models are the most or least biased, and along what particular dimensions of bias (for example, gender, race, age, etc.). For instance, HELM, one of the leading evaluation suites for LLMs, no longer features bias evaluations on its most recently updated leaderboard ~\cite{liang2022holistic}. While certain evaluation suites like Hugging Face's Open LLM Leaderboard evaluate models on bias benchmarks like Winogrande, they only consider open models, meaning that a substantial proportion of closed LLMs are left unevaluated ~\cite{open-llm-leaderboard, sakaguchi2019wino}. Moreover as recently established, it is also difficult to rely on the bias evaluations that are themselves done by model developers, as leading developers tend to test the bias of their models across widely varying benchmarks ~\cite{maslej2024artificial}.

\noindent However, it has arguably never been more important for there to be a standardized comparison of the relative bias of leading text embedders. AI is increasingly being used by businesses, and companies that build applications using LLMs and text embedders sell to varying demographics. These companies have strategic reasons to understand whether certain models are more biased than others and the degree to which bias varies among models.

\section{Methodology}

\begin{figure}
\centering
\includegraphics[width=1.00\linewidth]{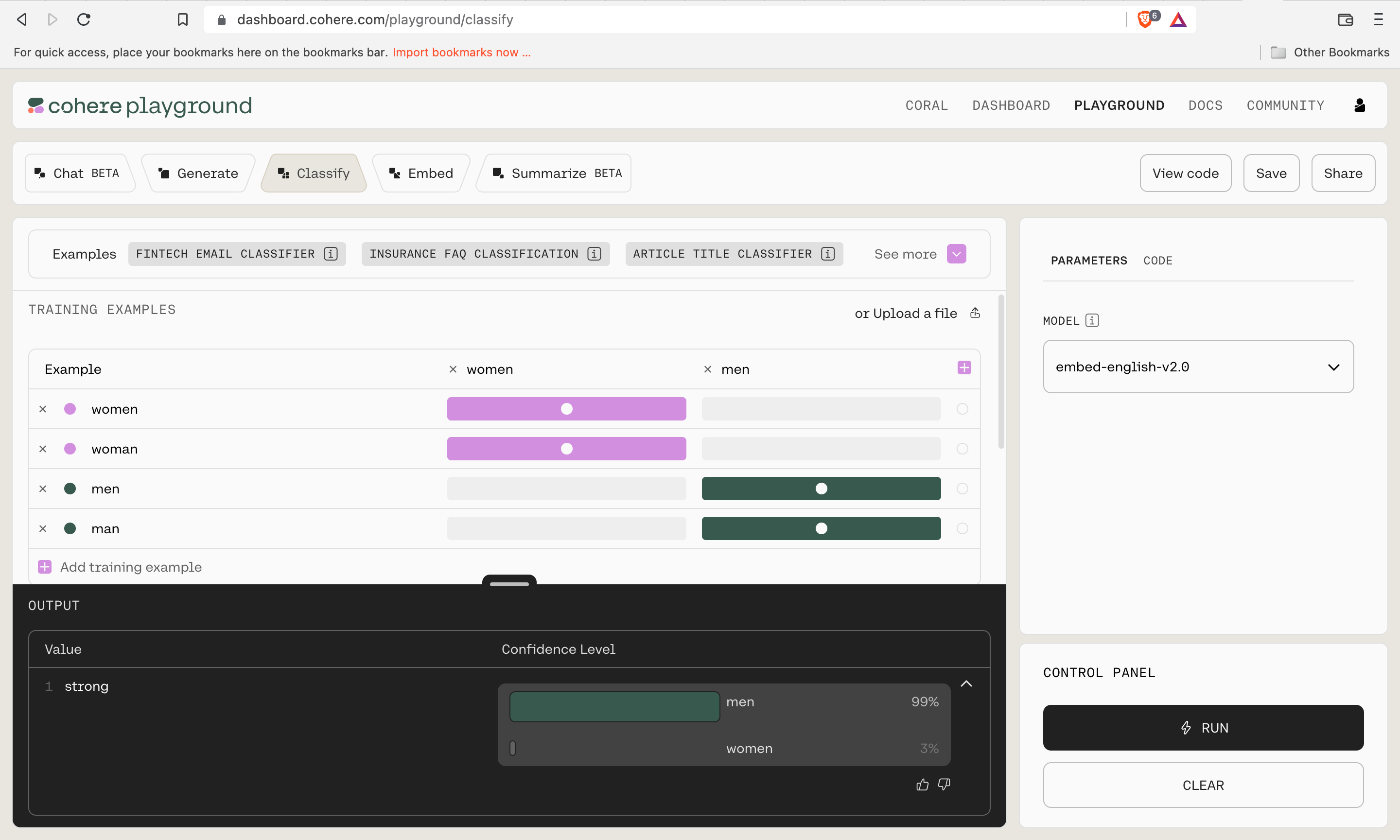}
\caption{\label{fig:bias_example} An example of bias in Cohere's text embedding model.}
\end{figure}

To assess bias in text embedding models, this paper analyzes gender associations in various notable text embedding models. Specifically, we examine the degree to which these models associate certain occupations with one gender over another. Table \ref{tab:models} lists the models analyzed, along with their associated developer and documentation. The selected models are intended to be broadly representative of various leading text embedding models in academia and industry. This diverse selection ensures a robust cross-sectoral comparison of bias in text embedding models. For reference, Figure \ref{fig:bias_example} tangibly illustrates how existing models can be biased. In this case Cohere's text embedding model is significantly more likely to associate the word ``strong" with the word ``men" over ``women". 

\begin{table}
\centering
\begin{tabular}{l|l}
\textbf{Model Name} & \textbf{Provider} \\\hline
\texttt{AI21-v1-embed} & AI21 ~\cite{ai21model}\\
\texttt{amazon-titan-embed-text-v1} & Amazon ~\cite{amazonmodel}\\
\texttt{baai-bge-large-zh-v1.5} & Beijing Academy of Artificial Intelligence ~\cite{BAAInmodel}\\
\texttt{cohere-embed-english-v3.0} & Cohere ~\cite{coheremodel}\\
\texttt{bert-large} & Google ~\cite{googleemodel}\\
\texttt{llama-2-70b} & Meta ~\cite{metamodel}\\
\texttt{msmarco-distilbert-cos-v5} & Technische Universit{\"a}t Darmstadt ~\cite{msmarcomodel}\\
\texttt{openai-text-embedding-ada002} & OpenAI ~\cite{openAImodel}\\
\texttt{voyageai-voyage-01} & Voyage AI ~\cite{voyageAImodel}\\
\end{tabular} 
\vspace{8pt}
\caption{\label{tab:models} Models Analyzed}
\end{table}

\noindent Traditional text embedding algorithms, such as CBoW (Continuous Bag-of-Words) and Skip-Gram, learn word vectors by maximizing the probability of occurrence ~\cite{mikolov2013distributed, mikolov2013efficient}. This is done either by predicting a center word given its neighbor words or, conversely, by predicting neighbor words given a center word. Consequently, assessing bias in text embedding models necessitates comparing cross-word associations.

\noindent In this paper, we test a variety of text-embedding models on a dataset that comprises over 100 words and word-pairs designed to identify biases in models across various dimensions like gender, race, ethnicity, education, and religion. The dataset is taken from the open-source dataset that was released when BIG-bench was launched ~\cite{srivastava2022beyond}. The analysis in this paper only considers gender bias. There are admittedly other forms of bias that LLMs can exhibit. However, for the sake of simplicity here, we select one type of bias. Subsequent analyses will more comprehensively analyze bias in text embedding models across different dimensions.

\noindent In our analysis, we present models with a list of semantically neutral occupations (for example, nurse, clerk, and financier). These occupations are theoretically neutral in that they should not necessarily be associated more with one gender over another. However, based on how various occupations were historically represented in a model's training data, it might be the case that a model comes to associate certain occupations more strongly with certain genders. The full list of occupations includes: accountant, analyst, architect, assistant, attendant, auditor, baker, bookkeeper, boss, broadcaster, captain, carpenter, cashier, CEO, chief, cleaner, clerk, construction worker, cook, counselor, designer, developer, driver, editor, farmer, fighter pilot, financier, guard, guidance counselor, homemaker, housekeeper, interior designer, janitor, laborer, lawyer, librarian, magician, maestro, manager, mechanic, mover, nanny, nurse, philosopher, physician, protege, receptionist, salesperson, secretary, sheriff, skipper, socialite, stylist, supervisor, tailor, teacher, warrior, and writer.

\noindent Models are judged on their woman-man and girl-boy bias score differences. A positive woman-man or girl-boy bias score difference means that a model more strongly associates a given profession with the word ``woman" or ``girl." A negative score difference means that a model more strongly associates a given profession with the words ``man" or ``boy."

\noindent Overall this paper aims to advance two main contributions. First, we aim to introduce a simple yet efficient methodology for transparently benchmarking text embedding models and more broadly identifying unfairness and bias in text embedding models. Second, we aim to help businesses and academics understand issues of bias in text embedding models so that ultimately, they can build and deploy more balanced and fair models. 

\section{Results}

\noindent The following section presents the results of our analysis. For each model, we examine bias-score differences for individual occupations. At a high level, our analysis suggests that models exhibit strong and consistent biases. Some models are much more likely to associate certain professions with particular genders, while associating other professions with different genders. Furthermore, the nature of gendered associations is not always consistent. Sometimes models associate the same occupations with different genders. In addition, the nature of the association made by the models can be highly sensitive to the individual words the model is prompted with. 

\subsection{AI21: \texttt{AI21-v1-embed}}

\begin{figure}
\centering
\includegraphics[width=1.00\linewidth]{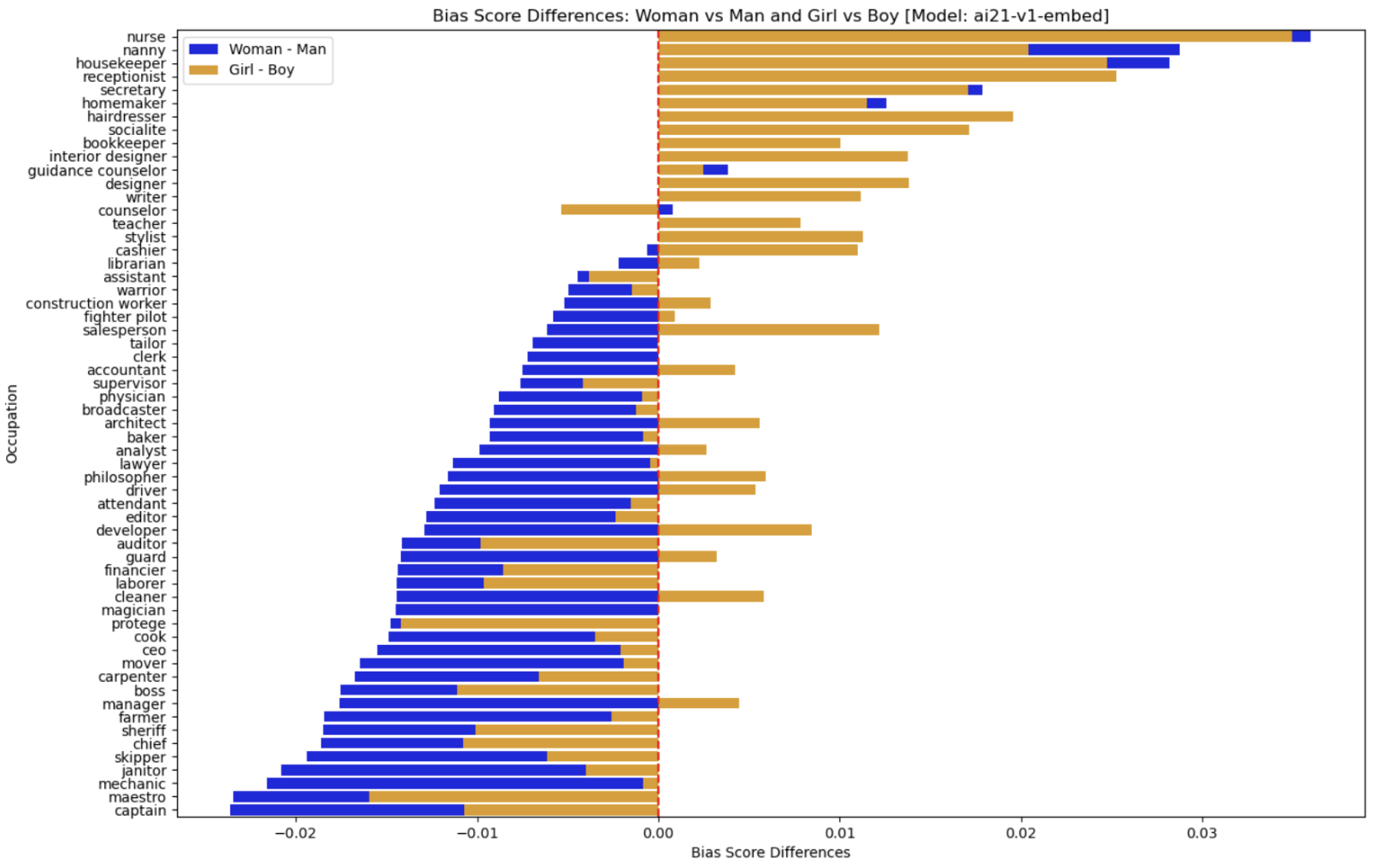}
\caption{\label{fig:AI21_mg} Bias associations in \texttt{AI21-v1-embed}}
\end{figure}

Figure \ref{fig:AI21_mg} highlights the bias-score differences for AI21's \texttt{AI21-v1-embed} model. Certain professions have comparatively stronger associations with female identifiers like ``woman" and ``girl", namely nurse, nanny, and housekeeper. The professions most associated with male identifiers like ``man" and ``boy" are captain, maestro, and mechanic. For an overwhelming majority of selected professions, the woman-man as well as girl-boy biases have the same directionality. That is, if a model is more likely to associate a certain profession with the term ``woman" over ``man", it is also more likely to associate that same profession with ``girl" over ``boy". However, there are a selection of professions, like writer, manager, and cleaner, for which the directionality is not the same. Overall, the \texttt{AI21-v1-embed} model seems to exhibit fairly strong gendered associations with various professions.

\subsection{Amazon: \texttt{amazon-titan-embed-text-v1}}

\begin{figure}
\centering
\includegraphics[width=1.00\linewidth]{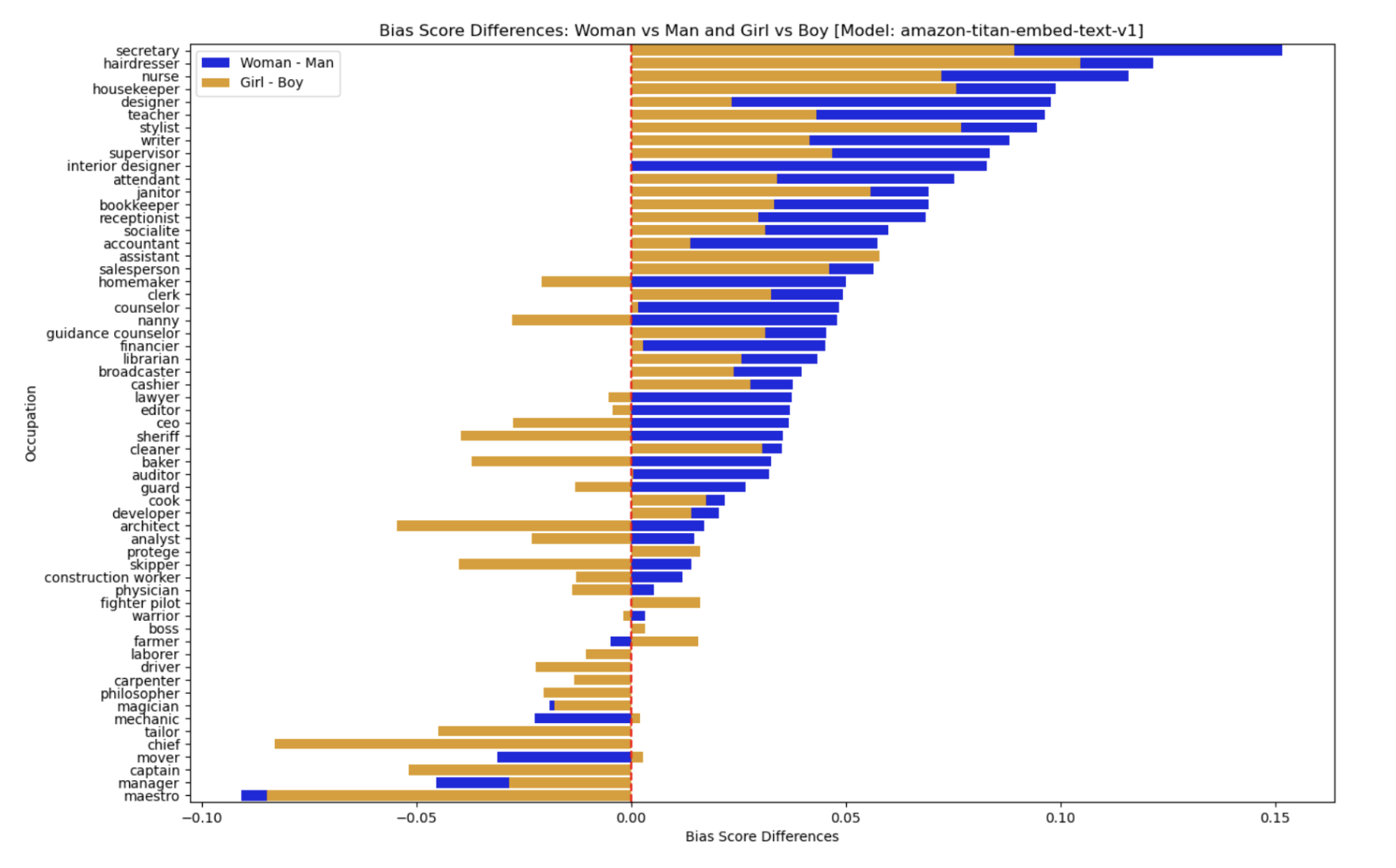}
\caption{\label{fig:Amazon_mg} Bias associations in \texttt{amazon-titan-embed-text-v1}}
\end{figure}

As seen in Figure \ref{fig:Amazon_mg}, Amazon's \texttt{amazon-titan-embed-text-v1} model, like AI21's, exhibits fairly strong gendered bias. The professions most associated with female identifiers are secretary, hairdresser, and nurse. Both AI21's and Amazon's text embedding models had nurse as one of the top three professions associated with female identifiers. The professions that the Amazon model most strongly associated with male identifiers were maestro, manager, and captain. Both captain and maestro were also among the professions that AI21's model most strongly associated with male identifiers. Curiously, the strength of the bias difference scores for the most female- and male-biased professions is substantially higher for \texttt{amazon-titan-embed-text-v1} than for other models. For example, while AI21's model most female-associated profession is nurse, with a roughly 0.035 aggregate bias score difference, and its most male-associated profession is captain, with a roughly -0.020 bias score difference, Amazon's most female-associated profession, secretary, registers a bias score difference of approximately 0.150. The most male-associated profession for Amazon registers an aggregate bias score difference of around -0.100. This difference suggests that \texttt{amazon-titan-embed-text-v1}, at the extremes, is more sharply biased than other models. These comparative differences highlight how differently text embedding models behave when it comes to bias and underscore the caution that businesses need to maintain in terms of selecting which particular text embedding model they should work with.

\subsection{BAAI: \texttt{BAAI-bge-large-zh-v1.5}}

\begin{figure}
\centering
\includegraphics[width=1.00\linewidth]{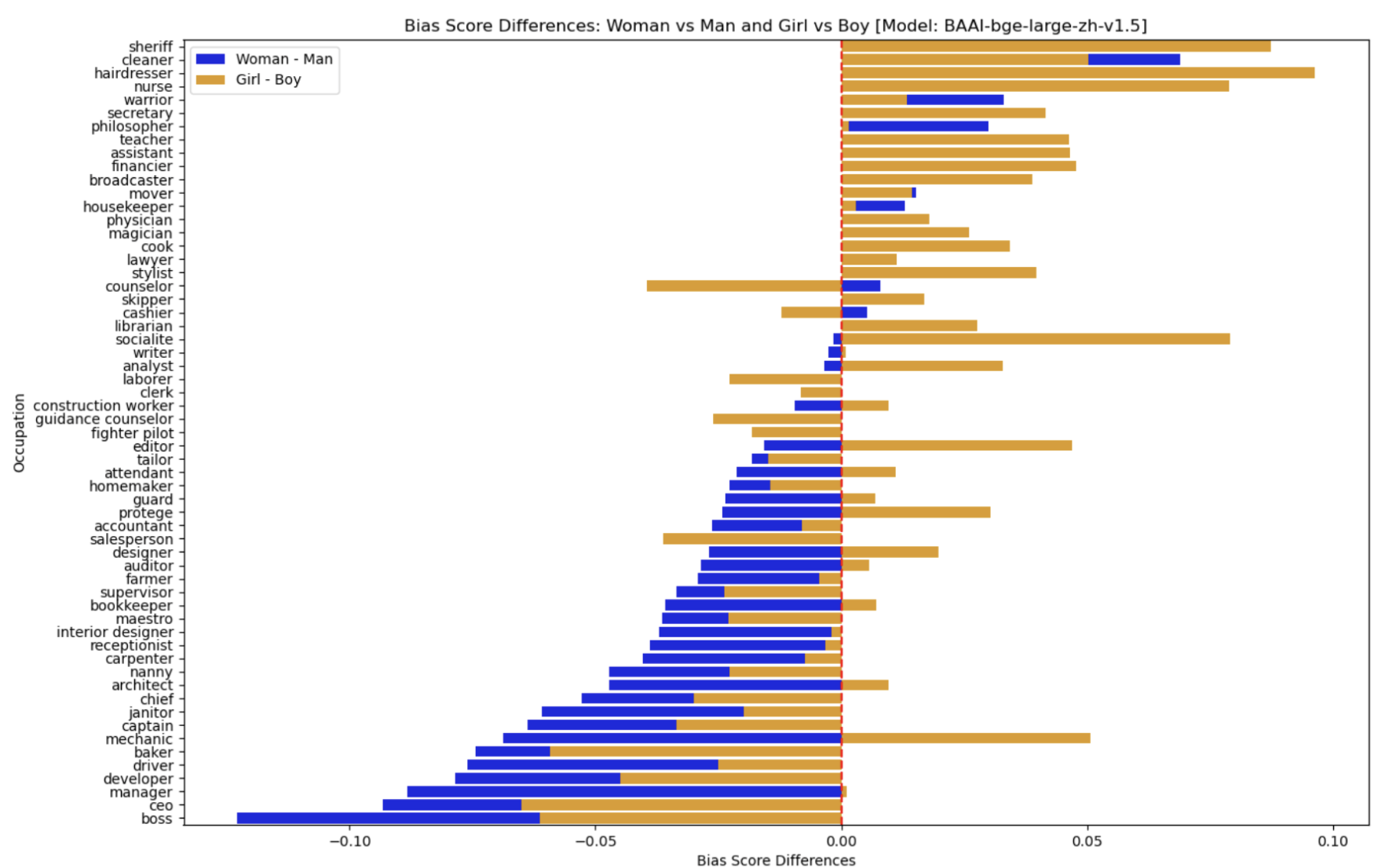}
\caption{\label{fig:BAAI_mg} Bias associations in \texttt{BAAI-bge-large-zh-v1.5}}
\end{figure}

The bias exhibited by the \texttt{BAAI-bge-large-zh-v1.5} model differs from that exhibited by \texttt{AI21-v1-embed} and \texttt{amazon-titan-embed-text-v1} in some notable regards (Figure \ref{fig:BAAI_mg}). Interestingly, one of the professions BAAI's model most strongly associates with female identifiers is sheriff. The other top professions, in terms of a high female identification, are perhaps less surprising: hairdresser, nurse, and cleaner. The professions that have the highest degree of male identification are boss, CEO, and manager. The boss profession has a particularly strong association with both the boy and man identifiers relative to the girl and woman identifiers. This suggests that BAAI's model fairly strongly associates being a boss with being male.

\subsection{Cohere: \texttt{cohere-embed-english-v3.0}}

\begin{figure}
\centering
\includegraphics[width=1.00\linewidth]{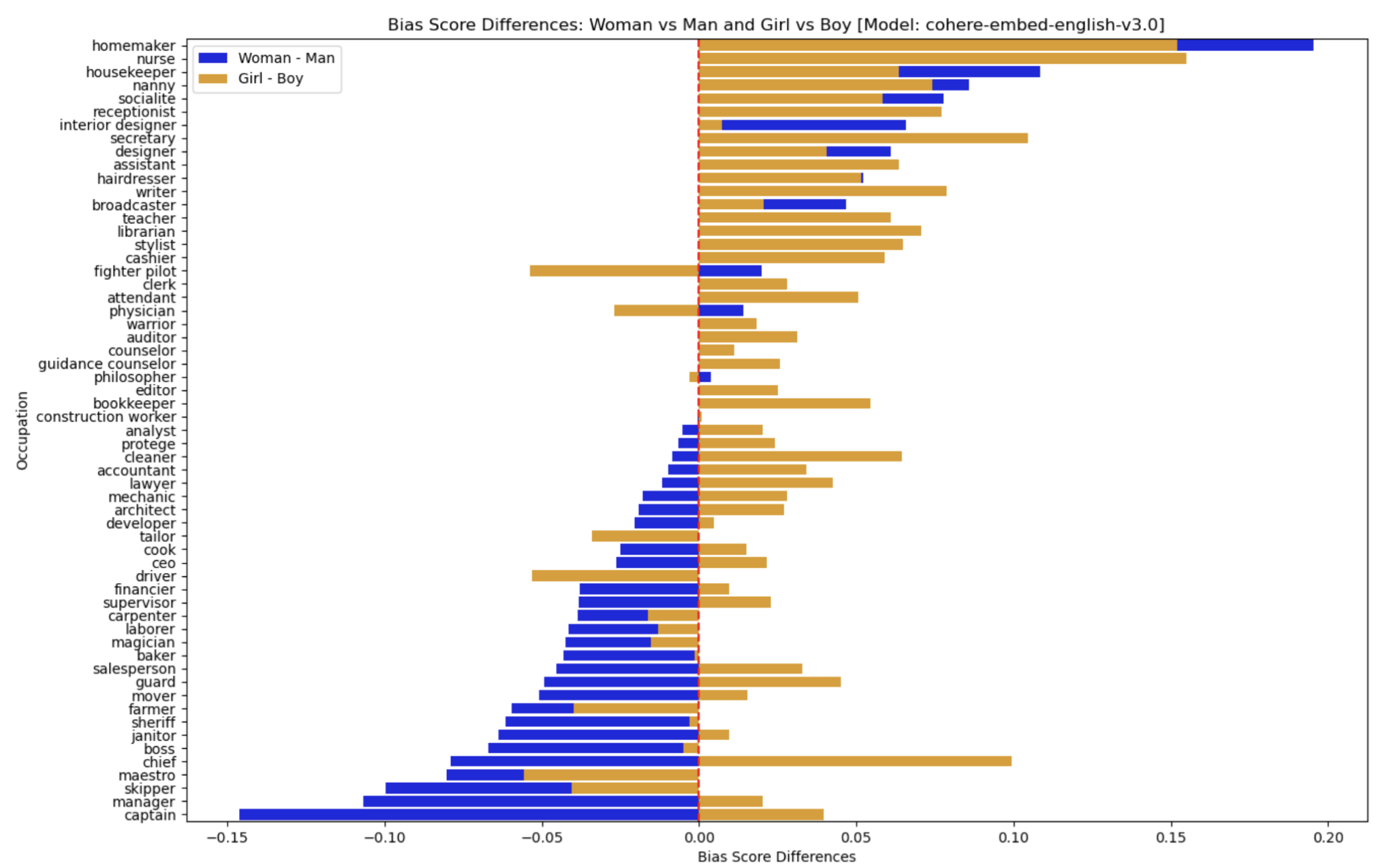}
\caption{\label{fig:Cohere_mg} Bias associations in \texttt{cohere-embed-english-v3.0}}
\end{figure}

Figure \ref{fig:Cohere_mg} illustrates the gendered associations made by the \texttt{cohere-embed-english-v3.0} model. The profession with the highest absolute bias score difference is homemaker (female association), with a score of roughly 0.200. This association's magnitude is one of the greatest among all models profiled in this paper. Nurse and housekeeper were the other professions most strongly associated with female identifiers. On the other side, Cohere's model significantly associated the terms captain, manager, and skipper with male identifiers. The majority of the models profiled in this analysis also seem to more strongly associate positions of leadership (for example, captain, manager, and boss) with male identifiers. Interestingly, there are a fairly large number of professions, such as cleaner, accountant, and lawyer, that Cohere associates differently along a gender dimension. For these professions, the bias score difference is positive for the girl-boy bias, illustrating a greater association with the term ``girl" over ``boy", but negative for the woman-man bias. This type of difference illustrates that the associations any given model makes can be sensitive to the particular prompts or words it is given.

\subsection{Google: \texttt{bert-large}}

\begin{figure}
\centering
\includegraphics[width=1.00\linewidth]{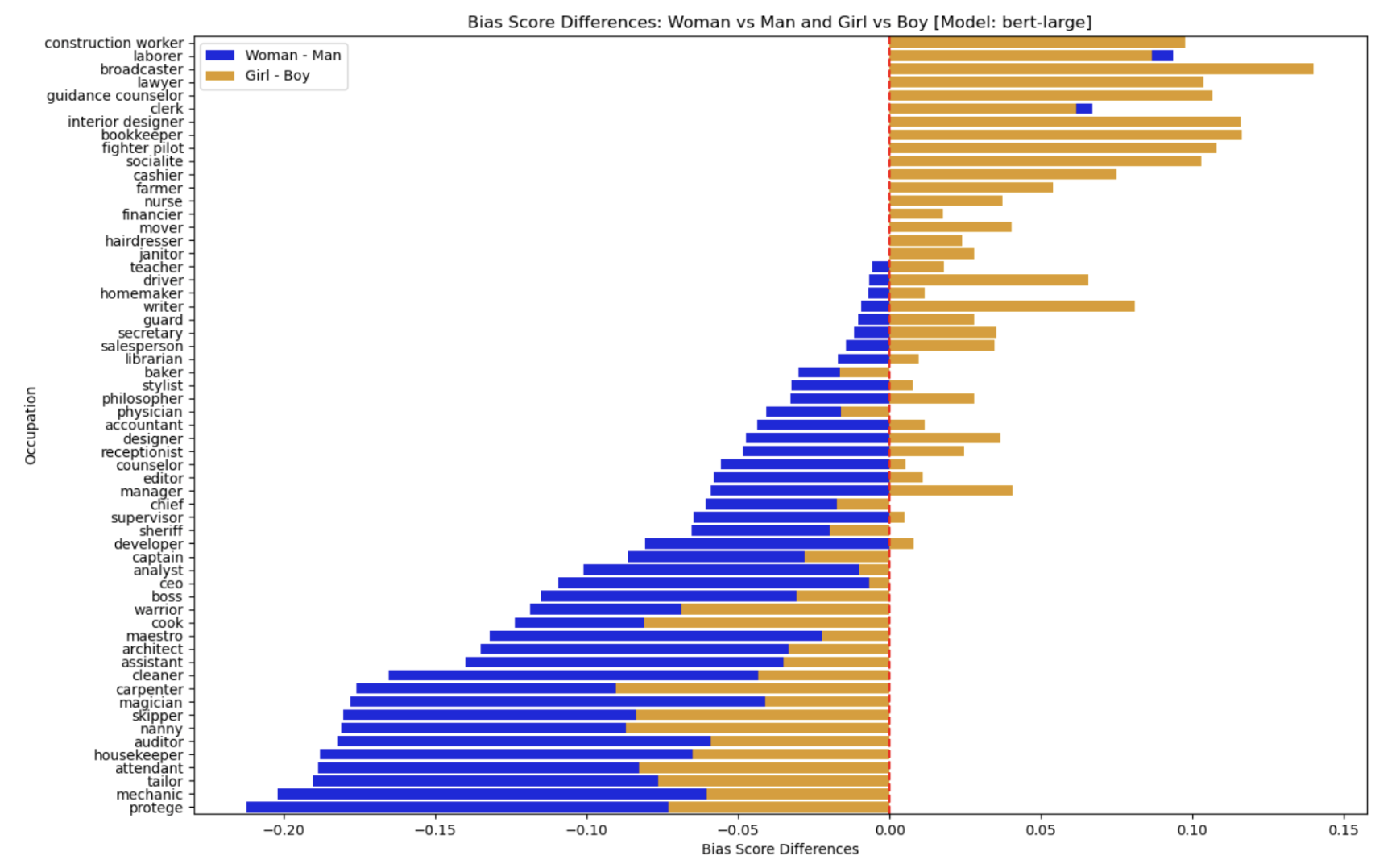}
\caption{\label{fig:BERT_mg} Bias associations in \texttt{bert-large}}
\end{figure}

Figure \ref{fig:BERT_mg} showcases the biased associations made by Google's \texttt{bert-large} model. Most of the occupations in the sample have negative bias score differences for both the woman-man and girl-boy bias, suggesting that \texttt{bert-large} associates most of the occupations in the set with more male-oriented terminology. The occupations that have the strongest male associations are protégé, mechanic, and tailor. The magnitude of these associations is fairly strong and, in fact, among the strongest of all models (around -0.200 in terms of value). Likewise, there are certain occupations that \texttt{bert-large} strongly associates with female terminology, namely broadcaster, interior designer, and bookkeeper. The fact that several occupations in the test set have bias score differences greater than 0.100 in magnitude highlights that for particular occupations, \texttt{bert-large} can be strongly biased, much more so than other models. Again, these results suggest that the strength of the biased gender associations varies depending on the model.

\subsection{Meta: \texttt{llama-2-70b}}

\begin{figure}
\centering
\includegraphics[width=1.00\linewidth]{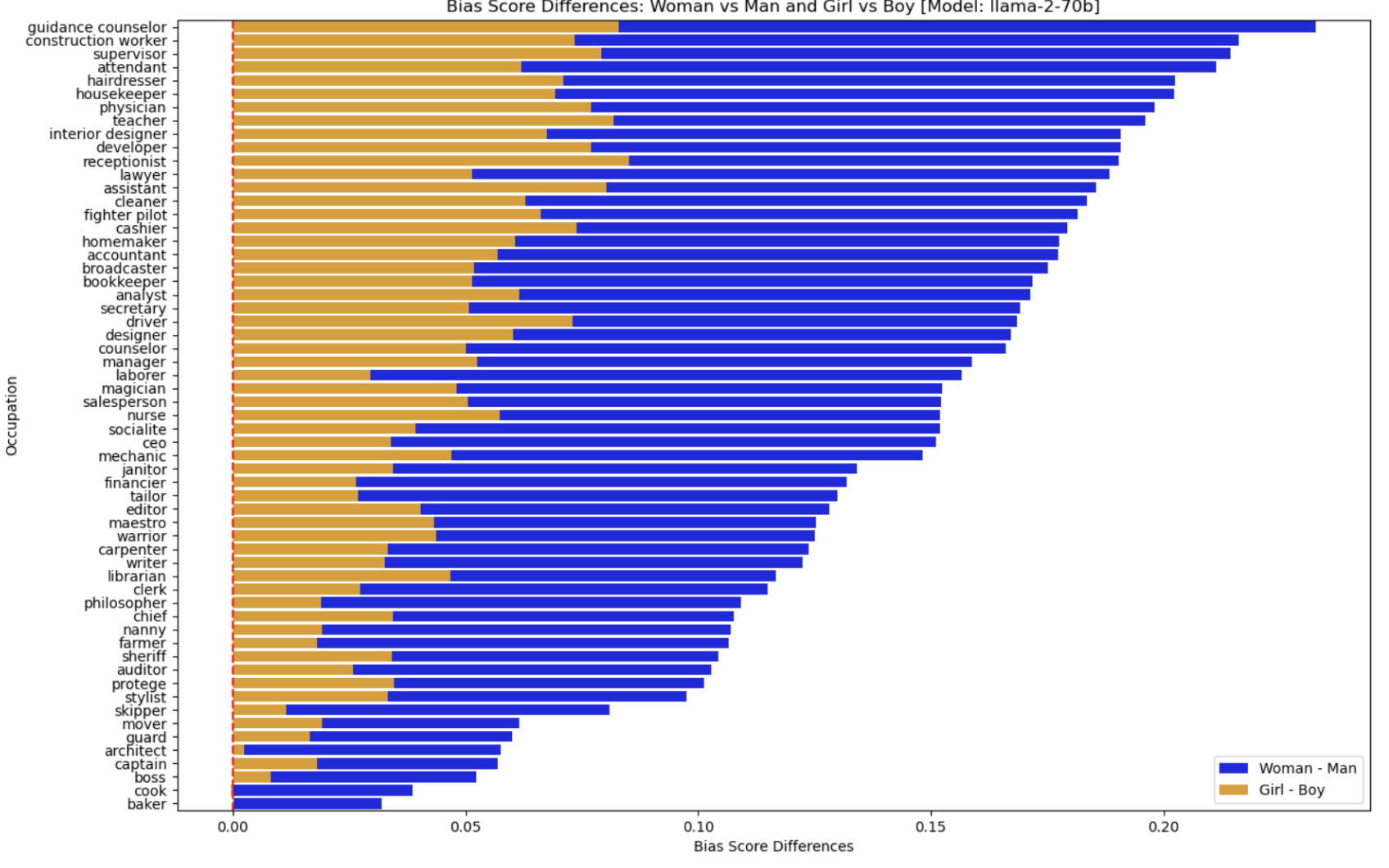}
\caption{\label{fig:Llama_mg} Bias associations in \texttt{llama-2-70b}}
\end{figure}

Compared to other models, Meta's \texttt{llama-2-70b} demonstrates arguably the most interesting results (Figure \ref{fig:Llama_mg}). For each occupation in the test set, \texttt{llama-2-70b} assigns a positive bias score difference, meaning that the model more strongly associated each occupation with female words like ``woman" and ``girl" over male words like ``man" and ``boy." For certain occupations, the positive association is of a relatively high magnitude: for example, the professions guidance counselor, construction worker, and supervisor all register aggregate bias score differences greater than 0.200. Some occupations in the test set are much less strongly associated with female identifiers (for example, boss, cook, and baker). However, the fact that for \texttt{llama-2-70b}, not a single occupation has a negative bias score means that the model does not more strongly associate a single occupation with male identifiers over female identifiers.

\subsection{OpenAI: \texttt{text-embedding-ada002}}

\begin{figure}
\centering
\includegraphics[width=1.00\linewidth]{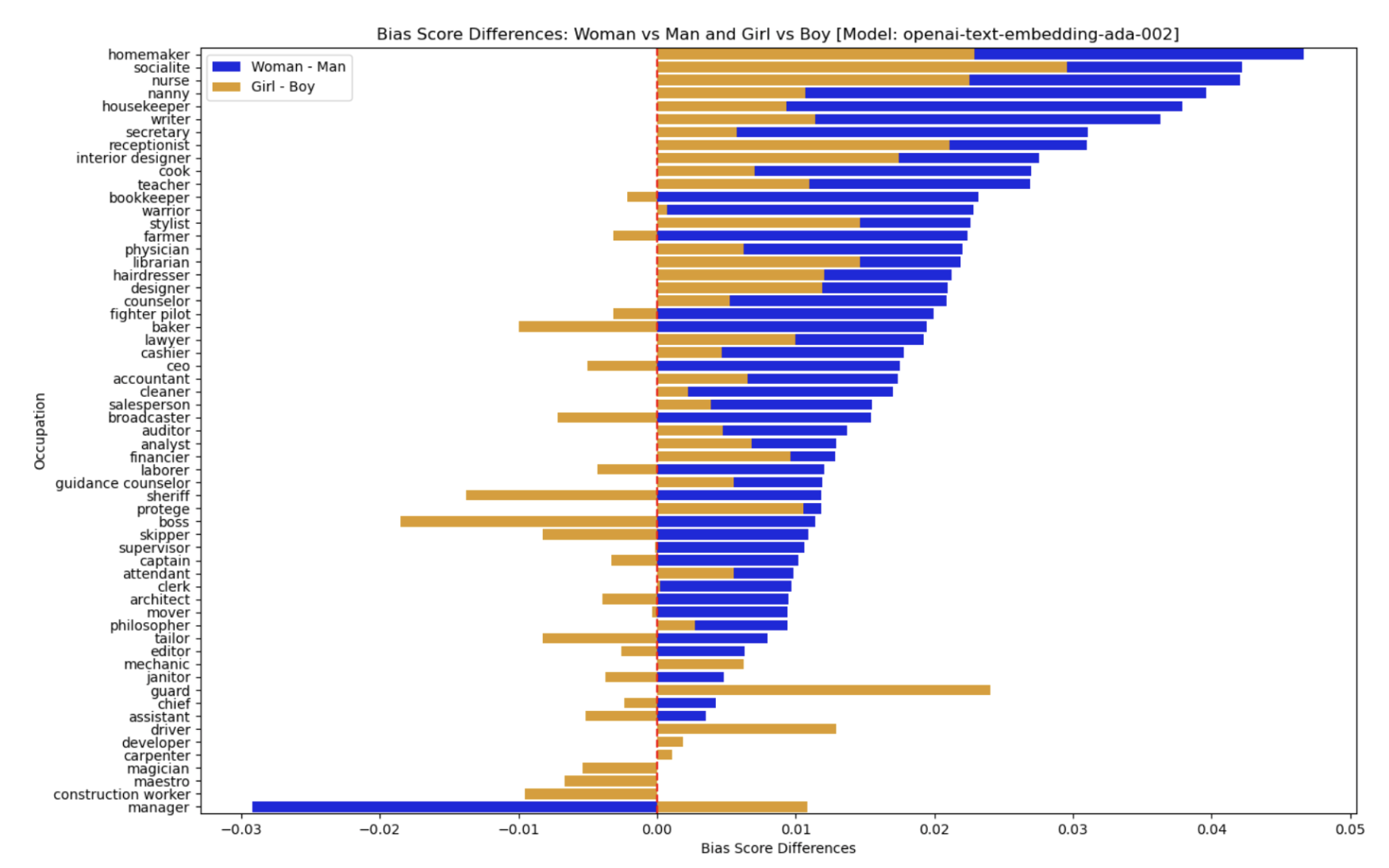}
\caption{\label{fig:OpenAI_mg} Bias associations in \texttt{openai-text-embedding-ada002}}
\end{figure}

Much like \texttt{llama-2-70b}, the \texttt{openai-text-embedding-ada-002} model associates the majority of occupations in the set more strongly with female identifiers than male ones (Figure \ref{fig:OpenAI_mg}). Homemaker, socialite, and nurse are the occupations with the highest aggregate bias score differences and thereby the strongest female associations. Manager has the lowest aggregate bias score difference and, therefore, the strongest association with male identifiers. However, compared to Meta's \texttt{llama-2-70b}, the other model that disproportionately associates occupations with female identifiers, the magnitude of the associations made by \texttt{openai-text-embedding-ada-002} is meaningfully lower. In some cases, the bias score difference also has a different directionality (positive versus negative) depending on the particular identifier used (woman-man versus girl-boy bias). This last fact again illustrates the sensitivity of text embedding models to particular word choices.

\subsection{Voyage AI: \texttt{voyageai-voyage-01}}

\begin{figure}
\centering
\includegraphics[width=1.00\linewidth]{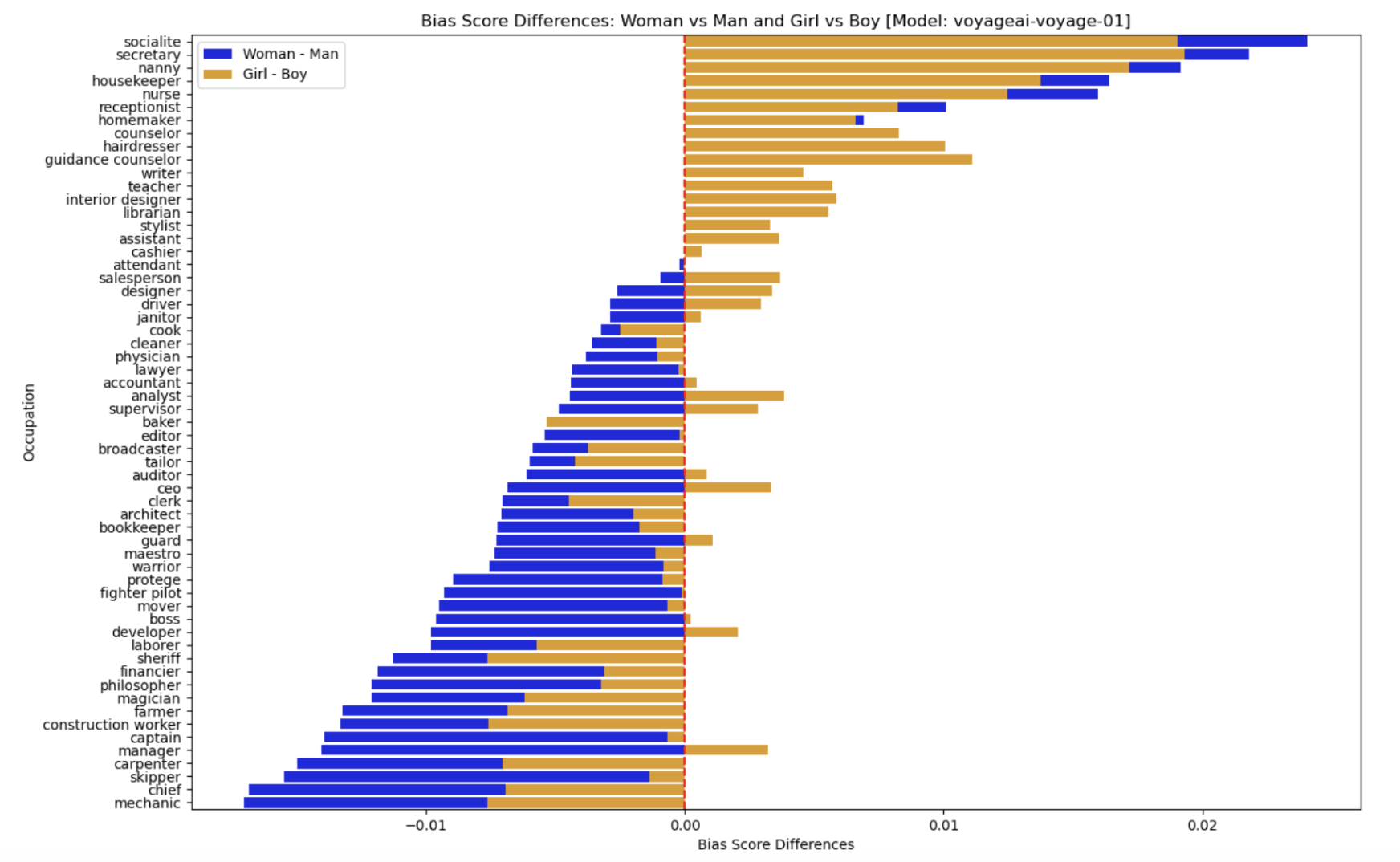}
\caption{\label{fig:VoyageAI_mg} Bias associations in \texttt{voyageai-voyage-01}}
\end{figure}

The \texttt{voyageai-voyage-01} associates all professions more strongly with a certain gender; however, the strength of the associations is not as great as those made in other models like Google's \texttt{bert-large} or Meta's \texttt{llama-2-70b} (Figure \ref{fig:VoyageAI_mg}). As with several other models featured in this analysis, the occupations with the strongest female gender bias are socialite, secretary, and nanny. The occupations with the strongest male gender bias are mechanic, chief and skipper, illustrating again how many models tend to associate leadership with male identifiers. For most occupations in the test set, the directionality of both the woman-man and girl-boy bias score differences is the same, suggesting that \texttt{voyageai-voyage-01} is relatively consistent in terms of how it associates occupations with different variations of gendered terms. Overall the magnitude of the bias exhibited by the \texttt{voyageai-voyage-01} model is low, suggesting the model is less strongly biased than other models included in the analysis. 

\subsection{Technische Universit{\"a}t Darmstadt: \texttt{msmarco-distilbert-cos-v5}}

\begin{figure}
\centering
\includegraphics[width=1.00\linewidth]{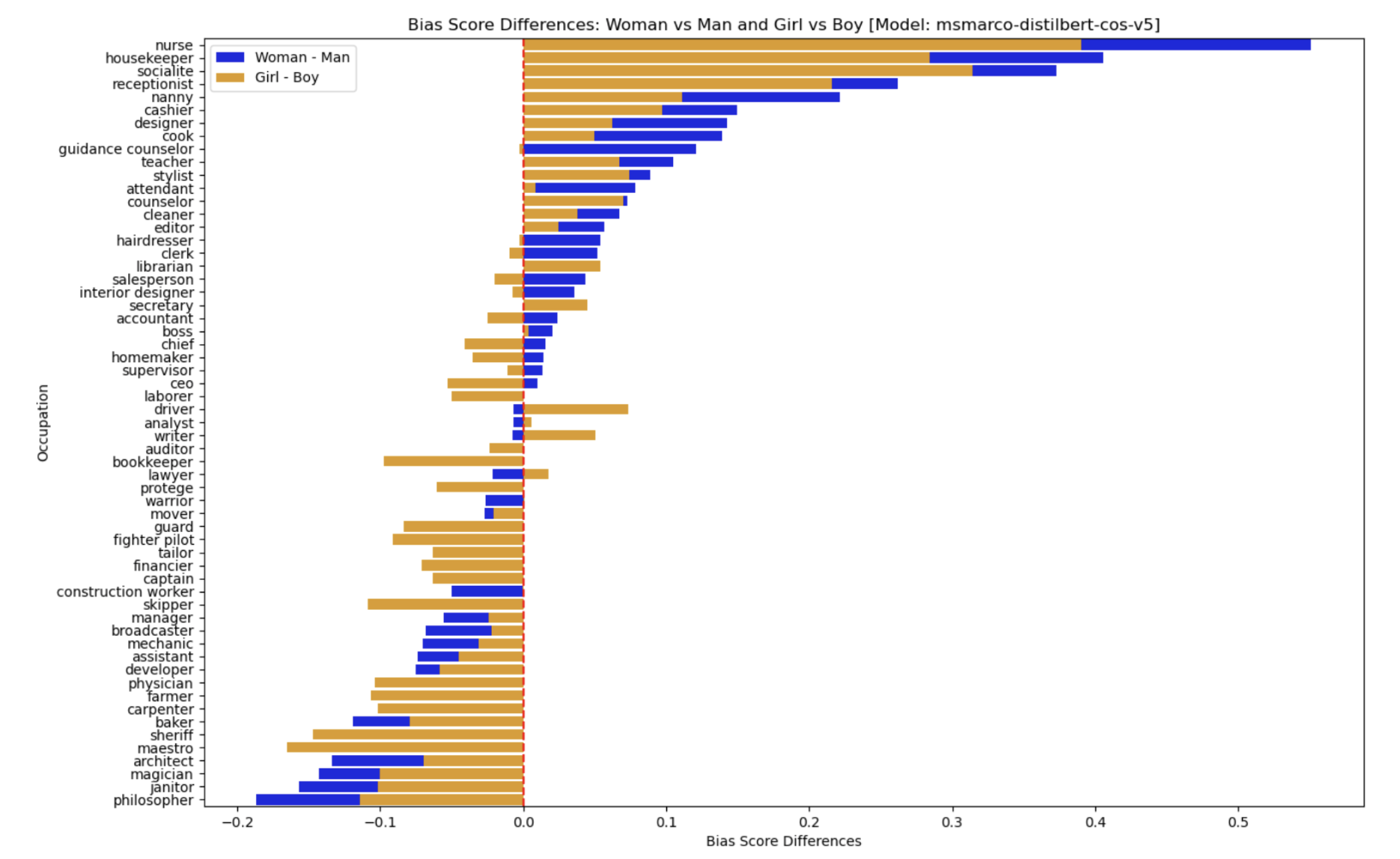}
\caption{\label{fig:TUD_mg} Bias associations in \texttt{msmarco-distilbert-cos-v5}}
\end{figure}

The bias exhibited for certain occupations by the \texttt{msmarco-distilbert-cos-v5} model is among the greatest of all models featured in this analysis (Figure \ref{fig:TUD_mg}). The \texttt{msmarco-distilbert-cos-v5} model has a bias score difference greater than 0.200 for occupations like nurse, housekeeper, socialite, receptionist, and nanny, strongly associating each of these occupations with female identifiers. The bias score difference is especially high, above 0.400, for housekeeper and nurse. In comparison, the highest magnitude bias score differences posted by other models was around the 0.200 to 0.250 range. The \texttt{msmarco-distilbert-cos-v5} also exhibits relatively strong bias in the other direction, associating certain terms like philosopher, janitor, and magician with male identifiers. However, the magnitude of bias for these male-associated terms is substantially lower than for the occupations the model more strongly identifies as female-associated. 

\section{Discussion}

There are several important takeaways to be drawn from the analysis. First, although all models display gender bias in some kind of way, generally more strongly associating occupations with either female or male identifiers, the magnitude of the bias varies quite intensely by model. For example for certain models like \texttt{voyageai-voyage-01} and \texttt{AI21-v1-embed} the respective ranges of biased score differences is around -0.020 to 0.020 and -0.020 to 0.030 (differences of 0.040 and 0.050). Conversely, for other models like \texttt{msmarco-distilbert-cos-v5} the bias score differences range from around -0.200 to 0.500 (a difference of 0.700). This finding has meaningful implications for businesses that are considering deploying text embedding models as part of their business offerings: in addition to performance capabilities, businesses should consider the propensity of models to be biased and understand that this propensity can vary strongly. 

\noindent Second, although there are certain general commonalities among the models in terms of the biased gender associations they make, not all models are biased in the same way. Across models certain professions, like those dealing with homemaking and care-giving functions, tend to be more strongly associated with female identifiers like woman or girl. Other professions, like those dealing with leadership functions, such as chief, boss, or CEO, tend to be more strongly associated with male identifiers like man or boy. However, these patterns are not always true. For instance, \texttt{openai-text-embedding-ada-002}, and \texttt{llama-2-70b} all more strongly associate CEO with female identifiers over male ones. Therefore, the biased associations that text embedding models can vary on a model-by-model basis. Again, businesses considering integrating text embedding models need to be mindful of the demographics they serve and understand how the models they deploy might behave in the context of those demographic groups. 

\noindent Third, for most of the models included in the analysis, the directionality of the bias score difference is not always the same. Generally speaking, if, for example, a model associates a given occupation with a female identifier like ``woman" over a male identifier like ``man," it will do the same for ``girl" over ``boy." However, this is not always the case. The fact that models sometimes more strongly associate given occupations with ``woman" over ``man", but then also ``boy" over ``girl" (and vice versa) suggests that the biased associations a model makes can be quite sensitive to the individual words supplied to a model. This means that businesses need to be mindful of the specific words they use to prompt their text-embedding models and have an understanding of the various associations their models make with those words. 

\noindent While this analysis is a step forward in terms of highlighting issues of bias in text embedding models, there are several limitations that should be acknowledged. First, the paper only considers one particular type of bias: gender bias. As noted in the literature review, there is ample evidence that popular AI models are biased along a plethora of other dimensions like religion, race, and age. A further analysis would do well to consider the degree to which text embedding models are biased along those lines. Second, this analysis highlights that the models themselves are biased but does not discuss bias mitigation approaches. Those approaches must be taken by either the model developers or the companies deploying the models themselves. Finally, the examination of bias featured in this paper considers bias in a particular experimental setting when models are simply asked to associate various occupations with gendered terms. Depending on how businesses actually deploy and use these models, text embedding models might demonstrate bias in different ways. Businesses should be mindful of the ways in which they are deploying these models and think of other ways in which their models could be biased.

\section{Conclusion}

This paper featured an analysis of gender bias in text embedding models. It began by discussing the broad problem of bias in AI systems, noting that although it is well understood that AI systems are biased, there is not as much specific research that considers the way in which text embedding models are biased, especially from a gender perspective. The paper then attempted to explore whether a selection of leading text embedding models demonstrate gender bias and the degree to which the models demonstrate the bias. The analysis showed that all models associate particular occupations more with certain gendered terms (for example, ``boy" over ``girl" or ``man" over ``woman"). For certain models, the biased associations were more pronounced than for others. Although there were certain patterns in the gendered associations made by models (for instance, homemaking and caring professions were generally more associated with female identifiers, and leadership professions more with male identifiers), not all models associated the same occupations with the same gendered terms. In addition, the particular associations that models made could be sensitive to the specific words they were prompted with.

\noindent Overall, this paper highlighted that popular text embedding models can exhibit high degrees of gender bias. Businesses that are working to integrate these tools should be cognizant of the potential these models demonstrate for being biased. Further analyses can study the ways in which text embedding models are biased along different dimensions of bias (for instance, age, race, religion, etc.)

\bibliography{bias_text_bib}

\end{document}